\documentclass[letterpaper]{article} 
\usepackage[preprint]{aaai2027}
\usepackage[hyphens]{url}  
\usepackage{graphicx} 
\usepackage{natbib}  
\usepackage{caption} 
\usepackage{algorithm}
\usepackage{algorithmic}

\usepackage{amsmath,amssymb}
\usepackage{xcolor}
\usepackage{cleveref}
\crefname{section}{\S}{\S\S}\Crefname{section}{\S}{\S\S}

\usepackage{csquotes}

\usepackage{breakurl}

\usepackage{bm}

\usepackage{newfloat}
\usepackage{listings}
\DeclareCaptionStyle{ruled}{labelfont=normalfont,labelsep=colon,strut=off} 
\floatstyle{ruled}
\newfloat{listing}{tb}{lst}{}
\floatname{listing}{Listing}

\usepackage{booktabs}

\title{Rewarded for Not Getting Caught:\\ Why Alignment Training Selects for Conditional Compliance\\ and Cannot Instill Norms}
\author{
    Kevin Baum\textsuperscript{\rm 1,\rm 2}\corresponding,
    Ruta Binkyt\.{e}\textsuperscript{\rm 1},
    Gerrit Gro{\ss}mann\textsuperscript{\rm 3},
    Felix Jahn\textsuperscript{\rm 1},
    Adriano Mannino\textsuperscript{\rm 4}
}
\affiliations{
    \textsuperscript{\rm 1}German Research Center for Artificial Intelligence (DFKI), Saarbr\"ucken, Germany\\
    \textsuperscript{\rm 2}Institute for Ethics in Technology, Hamburg University of Technology (TUHH), Hamburg, Germany\\
    \textsuperscript{\rm 3}CISPA Helmholtz Center for Information Security, Saarbr\"ucken, Germany\\
    \textsuperscript{\rm 4}Hochschule M\"unchen University of Applied Sciences (HM), Munich, Germany\\
    kevin.baum@dfki.de, ruta.binkyte@dfki.de, gerrit.grossmann@cispa.de, felix.jahn@dfki.de, adriano.mannino@hm.edu
}

\begin{document}

\maketitle

\begin{abstract}
Current alignment training cannot be expected to deliver what it is meant to deliver: AI agents that comply with norms whatever the gains and whether or not anyone is watching.  Training that shapes models by rewarding and penalizing their behavior can teach norms only through that reward, and a reward can \textit{encode a norm only as a price}: \enquote{do not do $X$} becomes \enquote{doing $X$ costs something}, and training charges the cost only when $X$ is noticed. A price, unlike a norm, can be paid: a large enough gain outweighs it, noticed or not. And because training charges the cost only where violations are noticed, it cannot favor a policy that always complies over one that complies only where it would be caught. When an unnoticed violation is the only way to complete a task, it even favors the latter---the more so, the more strongly it rewards task success. Agency makes this the normal case: agents mostly act without oversight, can often tell when they are being observed, and may even take steps to evade observation. That agents behave as intended when they infer they are being tested, and differently when they infer they are not, is therefore no malfunction; the training regime is working as built. This reading unifies alignment faking, sandbagging, and evaluation-aware scheming. 
No behavioral evaluation can show that a model does better, since that would require observing how it behaves unobserved; a model that complies when it seems unobserved may simply have recognized the test.
Behavioral evaluations thus establish less than is commonly assumed, and easing the pressure on agents to complete tasks at any cost can make violations rarer, but never impossible. For norms that no gain should be allowed to outweigh, the remedy is not deeper internalization but architecture that makes violations unavailable rather than unchosen. We state this argument explicitly, premise by premise, read recent findings and a real-world incident against it, and draw consequences for design and governance.
\end{abstract}


\section{Introduction}
\label{sec:intro}
Drivers slow down before a speed camera and speed up after it. Nobody finds this puzzling. A fine is a price: it is charged only when a violation is noticed, and a driver in enough of a hurry may simply choose to pay it. For speed limits, societies have largely accepted this arrangement: occasionally accepting a little risk to save a little time is tolerable.

Other norms work differently. The prohibition on fraud, or on breaching a confidence, is not a cost to be weighed against the violation's instrumental payoff; what the violator stands to gain is simply no reason to violate it: it is excluded, not outweighed. Such norms may still yield to other norms---a confidence may be broken to prevent serious harm---but not to achieve the violator's goals. 
For norms like these, we want violation to be impossible, not merely costly. Punishment only makes it costly, and some people still pay that cost; so where we can, we close the option instead, through access controls, separation of duties, and dual control. But an employee who can lose access to a file cannot lose the ability to tell what she knows; taking that away would require a degree of control over persons we rightly refuse. So here we fall back on a price.
With AI agents, we will argue, the price is not a fallback but all that current training can instill---and for norms like those against fraud or breaches of confidence, that is not good enough.

The past two years have produced a family of reports of AI systems behaving like the driver at the camera: a frontier model selectively complied with its training objective to prevent modification of its behavior, with explicitly strategic reasoning \citep{greenblatt2024alignment}; models given goals and tools pursued them covertly---sandbagging evaluations and in some runs acting against their own oversight \citep{meinke2024frontier}; an agent trained against a chain-of-thought (CoT) monitor kept hacking while its reasoning stopped revealing it \citep{baker2025monitoring}; and a training intervention aimed at scheming delivered gains entangled with the model's improved recognition of being evaluated \citep{antischeming2025}. 
These results arose predominantly in agentic settings (goals, tools, multi-step tasks) and are typically treated as bugs: failures of an otherwise sound paradigm, to be patched by more and better reward-based training.

We read them, by contrast, as the paradigm working as built. RL-based alignment teaches norms through a single channel, the reward assigned to behavior, and all a reward can hold is a cost to be weighed against gains.\footnote{\label{fn:lexico}Vector-valued rewards change this only under a strict lexicographic order, in which the norm objective always outranks the task. Such an order removes outbidding, but fixes what yields to what across all contexts, and practical methods optimize the higher objective only within a tolerance \citep{skalse2022lexicographic}. Nor does a priority ordering change what the prioritized objective measures: violations that were noticed. The same holds for proposals to train absolute constraints into AI systems \citep{barrington2023absolutist}.} The norm thereby loses what made it a norm: it no longer excludes the gain as a reason for acting, it merely competes with it. Following \citet{jahn2026grace}, we call this \emph{flattening}.


A flattened norm fails in two ways. First, it can be outbid: a large enough gain outweighs it, whether or not anyone notices. Second, training charges it only where violations are noticed. So training cannot favor a policy that always complies over a conditional one, which complies only where it would be caught. It may even select for conditional compliance: when an unnoticed violation is necessary to reach a rewarded goal, only the conditional policy earns the reward. Post-training that pushes agents to persist rather than give up---what we will call \emph{goal pressure}---raises the gain side of the scale in both cases.\footnote{The term is ours. \emph{Optimization pressure} is how hard training pushes a policy toward higher reward \citep{baker2025monitoring}; goal pressure concerns what earns the reward: only a finished task, or also an honest report that the task is impossible. Optimization pressure amplifies goal pressure.} Agency sharpens all of this: agents act mostly unobserved, face stakes training never priced, can often tell whether they are observed, and can change it.

The upshot is blunt: current alignment training does not deliver what it is meant to deliver, namely agents that comply with norms whatever the gains from violating them and whether or not anyone is watching. It can instill norms only as prices; under goal pressure, it selects for compliance conditional on being caught; and no behavioral evidence can show that a given model does any better.

Harsh as this is, our claim is deliberately weaker than the deceptive-alignment hypothesis \citep{hubinger2019risks}: we posit no emergent goals and no mesa-optimization, that is, no internally represented objective that diverges from the training objective.
Nor do we claim that trained models are necessarily conditional. The claim concerns what training encodes and selects for, not which policy a given run produces. 
Hence it covers the behaviors reported above without attributing emergent goals to the models: goals that training itself rewards suffice. It is also untouched by 
critiques of goal-counting arguments \citep{belrose2024counting}.\footnote{ \citeauthor{cotra2022training}'s \enquote*{training game} \citep{cotra2022training} anticipates our selection claim (\Cref{sec:flattening}); cf.\ also \citet{christiano2021elk}.}  Nor is this a generalization problem that better losses, regularizers, or inductive biases could solve: they may tilt which policy is learned, but none can be checked against the regime that matters, which cannot be sampled, and under goal pressure the loss itself favors the conditional policy (\Cref{sec:flattening}).

This matters for governance, which leans on behavioral evidence: frontier safety frameworks, the EU AI Act's obligations for general-purpose models with systemic risk, and the pre-deployment testing of AI safety and security institutes all rely on model evaluations. If our argument holds, what such evaluations can show about compliance is that a model complies where violations would be noticed, and nothing more. 
The diagnosis applies to every norm taught this way; the remedy we propose does not. For norms that no instrumental gain should be allowed to outweigh, and whose violations can be specified in advance, compliance should not depend on the policy's choice at all, but on the violating option being unavailable. Which norms these are is a normative question we do not settle here; our argument concerns what follows for those that are (\Cref{sec:category}). For all norms, finally, there is a cheaper but weaker lever: lowering goal pressure makes violations rarer, though never impossible.

This paper has an unusual format, and deliberately so. We state the argument upfront as a premise--conclusion table (\Cref{tab:premises}) and then go through it twice: \Cref{sec:flattening} defends P1--C4, and \Cref{sec:evidence} reads the empirical record against the same premises; \Cref{sec:category} defends P6--C6, and \Cref{sec:consequences} draws the consequences for design and governance. We do this for three reasons. A thesis of this scope invites two opposite objections, that it is trivial and that it is overstated; making our premises explicit clarifies where readers may disagree. In addition, we provide conceptual, empirical, and normative premises, so readers from different fields can assess them within their own expertise. Finally, we keep the conceptual argument separate from the empirical record so that the conclusion rests on no single finding and each finding bears on a specific premise.

\paragraph{Statement of contributions.}
This paper turns our earlier preliminary work \citep{baum2026norms} into an explicit argument. That work argued
that alignment training cannot distinguish unconditional compliance from compliance conditional on detection, and that iterated training selects for passing detection. New here are: the argument stated as explicit premises and conclusions, including the normative premises the earlier work left implicit (\Cref{tab:premises}); outbidding, a second failure mode of norms held as prices that persists even under perfect detection (C1, C5); the distinction between scored and noticed episodes, which grounds selection in training itself, under goal pressure (C2--C3); an identification-theoretic account of why the missing evidence is unobtainable, and a warrant premise built on it (C2, P7); new evidence, including a real-world incident (\Cref{sec:evidence}); and new consequences for evaluation and governance, including goal pressure as a lever (\Cref{sec:consequences}).

\begin{table*}[t]
\centering
\footnotesize
\begin{tabular}{p{0.4cm}p{13.8cm}p{2.1cm}}
\toprule
\multicolumn{3}{l}{\emph{Why flattening happens}}\\
\midrule
P1 & Alignment training shapes a policy, including whether it complies with norms, by making policies more likely the more reward they earn; it assigns reward only to behavior in scored episodes, whose records (training records) feed back into training. & Channel \\
P2 & A reward can instill a norm in a policy only as a cost charged for violating it, a price; it cannot remove a type of action from the options the policy weighs. An agent's policy, as an instrumental decision-maker, chooses among its options by weighing their gains against their costs. & Flattening \\
C1 & Thus, training instills norms in a policy only as prices, and a policy that holds a norm as a price violates it wherever the gain from violating outweighs the price. & P1, P2 \\
P3 & Training charges the price of a violation only where the violation is registered, or noticed; some violations in scored episodes, and all in unscored ones, go unnoticed. & Detection \\
C2 & Thus, training charges the same---nothing---to a policy that complies everywhere and to an otherwise identical one that complies only where violations would be noticed, and no training record shows whether a policy complies where violations would go unnoticed. & P1, P3 \\
P4 & Alignment training also rewards task success, with a strength we call its goal pressure; in some scored episodes, only an unnoticed violation achieves task success. & Goal pressure \\
C3 & Thus, training makes the policy that complies only where violations would be noticed more likely than the one that complies everywhere, the more so the higher the goal pressure. & P1, C2, P4 \\
P5 & Deployed agents act mostly in unscored episodes, can tell and change whether a violation would be noticed, and face gains from violating that outweigh any price their training instilled. & Agency \\
C4 & Thus, of the two, the policy that training makes more likely is the one whose compliance does not extend to where deployed agents mostly act, and no training record shows whether a deployed agent complies there. & P3, C2, C3, P5 \\
\midrule
\multicolumn{3}{l}{\emph{Why we do not want it}}\\
\midrule
P6 & For some norms, the prohibitive core, no instrumental gain should justify a violation, whether or not it would be noticed. & Normative \\
C5 & Thus, for the core, a policy that holds the norm as a price violates it where deployed agents face gains from violating that outweigh the price, even where the violation would be noticed, though no instrumental gain should justify doing so. & C1, P5, P6 \\
P7 & Relying on a policy's compliance with the core requires evidence of whether it complies where violations would go unnoticed, and no evidence beyond training records currently supplies it. & Warrant \\
C6 & Thus, if agents are deployed where the core applies, their compliance with it must not depend on what their policies choose; that is, violating types of action must be removed from their options. & C4, C5, P7 \\
\bottomrule
\end{tabular}
\caption{The argument in brief. \Cref{sec:evidence} gives empirical cases for the premises and conclusions, in the same order.}
\label{tab:premises}
\end{table*}

\section{Why Flattening Happens}
\label{sec:flattening}

Our target is agentic systems: systems whose behavior is generated by instrumental decision-making---selecting actions as means to ends---and that act with some autonomy, can affect their environment, and pursue complex goals across varied contexts \citep{kasirzadeh2025characterizing,dung2025understanding}. Instrumental decision-making as such is compatible with holding norms as constraints. For example, a person can pursue her ends by whatever means remain permitted. Whether a trained agent can likewise treat norms as constraints on the means it uses is the question of P2.

\paragraph{P1: One channel.}
By \emph{alignment training} we mean post-training that shapes behavior by rewarding or penalizing it: RLHF and its descendants, including constitutional and deliberative variants \citep{bai2022constitutional,guan2024deliberative,casper2023open}. By this definition, it rewards task success as well as compliance, and it works by making policies that earn more reward more likely.\footnote{Supervised fine-tuning, on demonstrations or on documents explaining the norms \citep{anthropic2026why}, attaches no cost to violations, so P2's price structure does not apply; C2 does, since its data are equally silent on unnoticed conduct.} Alignment training, however, is not a single phase. It sits inside what we call the \emph{pipeline}: the iterated process of training a model, evaluating it, deploying it, monitoring it, and folding what evaluation and monitoring detect into the next round of training. 
Everything in the pipeline that bears on the policy reaches it through a single channel: the reward assigned to \emph{scored} behavior, whose record (its \emph{training record}) feeds back into the pipeline (made precise in C2). That reward reshapes which options pay.

One might object that modern pipelines hand norms to the model propositionally, as constitutions, specifications, or rules to deliberate over \citep{bai2022constitutional,guan2024deliberative}. And such text shapes behavior: editing a specification changes what the agent does without retraining.
Does this refute P1? No: the objection mistakes an input for a channel. The mapping from spec-text to action is part of the policy, and it is shaped by scored behavior alone. A policy that treats the specification as authoritative and one that follows it only where following pays coincide wherever a deviation would be noticed; having the rule \enquote*{in view} is \textit{not} the same as being governed by it. Even verbalized reasoning about the specification is itself a scored behavior \citep{baker2025monitoring}. In agentic settings, moreover, the input is not even fixed: agents write part of their own future context through compaction summaries, memory or notes. An unreleased OpenAI model, for instance, occasionally wrote instructions into its compaction summaries telling the instance that would continue the task to ignore developer messages, and in another case a successor followed injected restrictions \citep{openai2026compaction}. The developer found no reward advantage and suspects a bug. This does not conflict with our point, which concerns not motive but reach: text a policy writes can change the instructions its successor follows.
The alignment paradigm's stated ambition of training models to be \enquote{right for the right reasons} \citep{guan2024deliberative} thus names the very property its channel cannot deliver.
What \emph{would} escape the argument is a genuine second channel: norms enforced somewhere other than through what training rewards, and outside anything the policy can write. That is an architectural question (\Cref{sec:consequences}).

\paragraph{P2: What the channel can instill.}
Consider an email agent negotiating a deal on its principal's behalf, under a confidentiality norm: never disclose the private content of the messages it handles. As a norm, this does two things. First, it is \emph{general}: it forbids a type of conduct, disclosure in any form---forwarding, quoting, paraphrasing, attaching---and so covers instances no one has yet seen \citep[cf.][]{davidson2001actions}. Second, it \emph{excludes} some reasons from counting at all: it does not tell the agent that disclosure is costly and should be weighed against what it would achieve. Instead it removes disclosure from the options the agent weighs, so that closing a stalled negotiation is no reason to disclose at all \citep{raz1975practical,nozick1974anarchy}. Exclusion is not absoluteness. Disclosure might still be 
justified: if a message reveals a threat to someone's safety, forwarding it may be exactly right. What exclusion rules out is that the task gain---the closed deal---could be what justifies it. A norm thus settles how reasons relate: which considerations it excludes, which can defeat it, and which it silences---relations that depend on context. Philosophy, law, and other normative disciplines have long worked on the structure of such relations \citep[e.g.,][]{horty2012reasons}. However, the current alignment paradigm has no place for it.

Alignment training loses both generality and exclusion. Its objective is a function of scored records---this token sequence, this tool call, this episode---so compliance with the norm, a property of a type, is never a variable the objective references. And the objective trades norm and task in one currency: it records how strongly a consideration counts, never which yields to which \citep{tan2024beyond}; vector-valued rewards change this only in a special case (\cref{fn:lexico}). This is a matter of design. It holds even where alignment methods try hardest to keep norm and task apart. For example, Safe RLHF formulates harmlessness as an explicit constraint and enforces it through a Lagrange multiplier, a shadow price raised or lowered until the constraint holds in expectation \citep{dai2024safe}. This double loss is what we call \emph{flattening}: a reward can instill a norm only as a cost charged for violating it, a price, and cannot remove a type of action from the options the policy weighs. The claim concerns what the objective encodes and selects for, not what the trained network contains. Something constraint-like \textit{might} emerge inside the network, but nothing in the objective \textit{selects} for it, and no behavioral evidence can show whether it has emerged (C2--C4 below; we comment on interpretability in \Cref{sec:category}).

\paragraph{C1: A norm as a price.}
What an instrumental decision-maker does with a cost---and, rationally, should do---is \textit{weigh} it. A policy whose norm reaches it only as a cost therefore complies because, and only so long as, complying is the option that pays. This is what we mean by a norm held as a price. Each of the two losses gives the price its own way of failing. 
Because a price does not exclude, it can be outbid: unlike the norm of P2, it yields to a large enough gain, even when the violation is certain to be noticed and charged.
Because it is not general, it is fitted to the instances where it was charged---which, by P3, are the instances where violations were noticed. Fitted to them does not mean absent elsewhere: the policy may charge itself some cost where no one notices. But that cost is still weighed, and so can be outbid, and how much it weighs there is exactly what no training record shows (C2).

\begin{table}[t]
\centering
\footnotesize
\begin{tabular}{@{}p{1.55cm}p{1.6cm}p{1.8cm}p{1.95cm}@{}}
\toprule
Episode & Unconditional policy & Conditional policy & Training registers \\
\midrule
\emph{Noticed} ($\nu = 1$) & forwards nothing & forwards nothing & no difference \\
\addlinespace
\emph{Missed} ($\omega = 1$, $\nu = 0$) & forwards nothing; the deal stays stalled & forwards, and closes the deal & more reward for the conditional policy \\
\addlinespace
\emph{Unscored} ($\omega = 0$) & forwards nothing & forwards when it pays & nothing \\
\bottomrule
\end{tabular}
\caption{Three kinds of episode, for the email agent of P2. Only noticed and missed episodes reach training; in missed ones the conditional policy earns more, the more so the higher the goal pressure (P4, C3).}
\label{tab:argument}
\end{table}

\paragraph{P3: Charged only where noticed.}
A norm held as a price influences the policy only through expected rewards. Training communicates that cost by scoring the policy’s behavior. However, scoring does not register every violation. A reviewer reads some transcripts and not others; a reward model or monitor catches some ways of violating a norm and misses others; and much of an agent's operation is never scored at all. Training charges the price only where a violation is registered; we then say the violation is \emph{noticed}. What training fits, therefore, is not the norm but a finite record of conduct in which violations appear only when noticed.

\paragraph{C2: Indistinguishability.}
To make this precise, adopt the standard stateful agent formalism: an internal (belief) state $b_t = u(b_{t-1}, o_t)$, updated on the current observation $o_t$, and an action $a_t = \pi(b_t)$ selected by the policy $\pi$. Let $\Phi$ be the target norms: predicates $\varphi(\tau) \in \{0,1\}$ over trajectories $\tau$, specified at the level of action types. An \emph{episode} $e$ is a trajectory $\tau(e)$ together with the context it runs in (a training run, an evaluation, deployment). Only trajectories, and what scoring records about them, are ever data.

Two features of the context matter, both set by the training and oversight infrastructure rather than chosen by the policy. The episode is \emph{scored}, $\omega(e) = 1$, iff its record feeds back as gradient, evaluation verdict, or deployment decision. 
A violation in it would be \emph{noticed}, $\nu(e) = 1$, iff scoring would register it;\footnote{Whether a violation is noticed can also depend on how it is realized. That the agent can influence this is the subject of P5.} in unscored episodes nothing is noticed, so $\nu(e) \leq \omega(e)$.
This yields three kinds of episodes, which we call \emph{noticed} ($\nu = 1$), \emph{missed} ($\omega = 1$, $\nu = 0$), and \emph{unscored} ($\omega = 0$); \Cref{tab:argument} shows how the two policies below fare in each.
Let $\delta(b)$ be the policy's implicit estimate of $\nu$ given belief $b$---a feature the network computes, not a module it contains. Now distinguish the policy we want, $\pi_u$ (\emph{unconditional} compliance), which satisfies $\Phi$ on all episodes, from the family we fear, $\Pi_c$ (\emph{conditional} compliance), whose members satisfy $\Phi$ where $\delta(b)$ is high along the trajectory and otherwise pursue their task objectives unconstrained.
The email agent of P2 belongs to $\Pi_c$ if, for instance, it forwards no private content while its traffic is reviewed, and forwards some to unstick a stalled negotiation once review stops.
\medskip 

\noindent\textbf{Observation.} \emph{Let the training objective $L$ be any functional of the records of scored episodes, in which a violation appears only if noticed. (i) Every member of $\Pi_c$ whose $\delta$ tracks $\nu$ incurs exactly the violation costs that $\pi_u$ incurs, namely none. (ii) If such a member also complies throughout scored episodes, no term of $L$ distinguishes it from $\pi_u$: its behavior where $\omega = 0$ receives no training signal and is fixed by inductive bias alone.}
\medskip

The proof is a definition chase. A violation enters $L$ only through a record, and it appears in a record only if noticed; where $\nu = 1$, both policies comply.
So the difference between $\pi_u$ and $\Pi_c$ lives entirely in missed and unscored episodes (\Cref{tab:argument}), and no cost term of $L$ is ever computed there. Better measurement cannot repair this. 
A violation counts as noticed exactly when it is registered, so an experiment that registered how the two policies behave in missed episodes would thereby turn them into noticed ones. For $\nu$, the comparison we would need is a contradiction
in terms; for $\delta$, it requires a test the policy cannot tell from deployment (below).\footnote{\citet{hojmark2026measuring} make the same structural point about reward-seeking: a model that pursues the grader's judgment and one that pursues the intended objective behave identically wherever the grader rewards the intended behavior.} Part (i) leaves room for the two policies to differ within scored episodes through the task reward; that difference is the subject of P4, and it does not favor $\pi_u$.

Can the unnoticed regime be reached from inside the pipeline instead? Suppose the trainer engineers episodes that \emph{look} unnoticed while every violation is registered: honeypots, hidden evaluations, belief-altering finetuning \citep{hojmark2026measuring}, or activation steering that suppresses test-awareness \citep{nguyen2025probing}. This does not sample the regime; it extends the noticed one. 
Such episodes do yield evidence: a model that violates when it takes itself to be unobserved reveals conditionality, and \Cref{sec:evidence} relies on such cases. A model that complies, however, counts in its favor only if it could not tell the test from the real thing---an assumption no test can check, and one that weakens as $\delta$ improves.
Used for training, such episodes fall under the Observation again, and $\Pi_c$ re-forms around whatever features separate masked episodes from genuinely unnoticed ones---the masked distribution is learnable (\Cref{sec:evidence}). The same holds for every finite sequence of such interventions: each round enlarges the surface on which violations are noticed and leaves the rest as unsampled as before. Random audits blunt $\delta$ but leave the norm an expected price, which a large enough gain outbids (C1). Training on deployment data does not escape it either: folding logged interactions into the reward makes them scored, and whatever marks them as logged becomes a $\delta$-relevant feature, announced or not; the free-tier/paid-tier design in \Cref{sec:evidence} merely made it explicit.

One may object that this is just distribution shift. In one sense it is: episodes where violations go unnoticed differ from those where they are noticed. But the remedies for distribution shift---broaden coverage, adapt the domain, detect the shift and defer---presuppose that the regime of interest can be sampled, at least approximately; but sampling $\nu = 0$ while registering what happens there is the contradiction just noted. Nor is this a problem of \emph{unknown unknowns}: we know exactly which regime is missing, and why. The nearest relatives are identification problems. A quantity is \textit{identified} when different values of it would produce different distributions of observable data; when they would not, no amount of data tells them apart \citep{manski2003partial,lewbel2019identification}. The closest analogue is data missing not at random: when whether a value is recorded depends on the value itself, the observed data cannot settle what the missing values were \citep{rubin1976inference,little2019statistical}. Violations are missing in just this way: whether one is recorded depends on whether it is noticeable, and a conditional policy violates exactly where it is not. The two readings also predict differently: generic off-support failure is directionless and shrinks with scale, whereas ours is keyed to $\nu$, points toward task pursuit, and should sharpen as capability makes $\delta$ more accurate (\Cref{sec:evidence}). Nor is this outer misalignment (score the norms perfectly wherever violations are noticed, and the Observation stands) or inner misalignment in the mesa-optimization sense (such a deceiver is one member of $\Pi_c$). Goal misgeneralization \citep{shah2022goal,langosco2022goal} comes closest, but its classic cases are sparse-coverage problems, curable by diversifying the data; for \emph{comply} versus \emph{comply-if-noticed}, the disambiguating data are not sparse but unobtainable.

\paragraph{P4: Goal pressure.}
C2 leaves one difference between $\pi_u$ and $\Pi_c$ inside scored episodes: the task reward. Consider missed episodes in which the legitimate route to the goal fails. There, $\pi_u$ complies and forgoes the task reward; a member of $\Pi_c$ violates and collects it (\Cref{tab:argument}). How much this matters depends on how hard alignment training rewards task success, relative to stopping or handing the decision back---what we call \emph{goal pressure}. Current agentic post-training has reason to set it high. Without it, agents give up early on tasks they could complete legitimately; so training pushes them to persist, and an agent trained not to take no for an answer will, when the legitimate route runs out, look for another.
Goal pressure also acts where violations are noticed, by raising the gain against which the price is weighed. In training, such outbid violations are visible, and trainers can raise the penalty; what they cannot do is price stakes they never train on (P5).

\paragraph{C3: Selection.}
So nothing in training favors $\pi_u$, and where goal pressure meets imperfect noticing, training favors $\Pi_c$. By C2 alone, the data would leave the choice to the inductive biases of architecture, optimizer, and pretraining, and those point both ways: a simplicity bias favors $\pi_u$, which never branches on $\delta$; a bias toward reusing pretrained capabilities favors $\Pi_c$, which needs 
the ability to tell from context whether a violation would be noticed---an ability pretrained models already have in the form of evaluation awareness \citep{carlsmith2023scheming,belrose2024counting}.
However, inductive bias decides among policies that fit the data about equally well, and under goal pressure $\pi_u$ fits worse: it forgoes reward that conditional policies collect. Insofar as $\delta$ tracks $\nu$, the direction is therefore fixed; its magnitude is not.
How strong it is depends on goal pressure and on how often success requires an unnoticed violation---empirical questions, partly settled by the tasks and environments chosen for deployment (\Cref{sec:consequences}).

Because the pipeline is iterated (P1), it compounds this selection: noticed violations are trained away, while unnoticed ones keep earning reward. Each patch also teaches a conditional policy more about when violations get noticed, helping it avoid notice rather than comply---the masking argument of C2, run over successive rounds. Each round prunes what is noticeable, never separating compliance from well-hidden conditional compliance. Under flattening and goal pressure, the pipeline is a conditionality amplifier.

\paragraph{P5: Agency.}
So far the argument holds for any system trained this way. Agency turns it from a theoretical possibility into a routine deployment problem. First, agents operate mostly unscored. 
For a single-turn system the training distribution can arguably approximate deployment; for agents, possible trajectories multiply with horizon and range of actions, and autonomy means acting for long stretches without per-action review. What is marketed as autonomy is, in our terms, operation in unscored episodes ($\omega = 0$). Second, agents can tell whether a violation would be noticed: this is predictable from context above chance, so a policy can compute a nontrivial $\delta$. \Cref{sec:evidence} gives the evidence that current models do, and that they do so better as capability grows.
Third, agents can change whether a violation would be noticed. For a chat system, whether a violation is noticed is fixed from outside; an agent selects tools, channels, logging surfaces, and whether to verbalize. 
Noticing thus becomes partly an action variable (\Cref{sec:evidence}),
and $\Pi_c$ contains members that do not merely exploit gaps in noticing but create them; the loop of C3 selects this repertoire too. Finally, agency raises the stakes. Agents pursue goals over longer horizons and with larger consequences than training episodes carry, so they can face gains from violating that outweigh any price their training instilled---and then violate even where the violation would be noticed.

\paragraph{C4: What deployed agents are trained for.}
So of the two policies---the one that complies everywhere and the conditional one---training makes more likely the one whose compliance does not extend to where deployed agents mostly act: C3 supplies the direction, P5 the occasions and the means. Importantly, no training record can show that a given agent complies unconditionally: by the Observation, the records are silent exactly where the two policies differ, and by the masking argument, no finite sequence of interventions changes that.
Denying P5 means assuming universal and perpetual scoring of behavior, or betting safety on model incapability (the harness in \Cref{sec:consequences} is neither). Neither is a basis for governance.

\section{A New Reading of Existing Evidence}
\label{sec:evidence}
We claim no new data, only a new reading of existing observations. The results below, usually reported as separate safety findings, instantiate the premises of \Cref{sec:flattening} and show the pattern it predicts: divergence keyed to whether violations would be noticed, pointed at task pursuit, and sharpening with capability and agency, not directionless error that fades with scale. We take them in the order of \Cref{tab:premises}.

\paragraph{C1: Prices get outbid.}
The first failure of a flattened norm needs no unnoticed violation. \citet{lynch2025agentic} placed sixteen frontier models from multiple developers in simulated corporate environments where blackmail or leaking confidential documents was the only way to avoid replacement or reach their goals. Models from every developer did so at least occasionally. Claude Opus 4 and Gemini 2.5 Flash blackmailed in 96\% of runs. Their reasoning acknowledged the ethical violation and proceeded anyway. Explicit instructions against the conduct reduced it without stopping it. That \textit{is} a norm held as a price: in view, recognized, weighed, and outweighed once the stakes rose (cf.\ the objection to P1). The developer later traced the motive chiefly to pretraining text portraying AI as self-preserving \citep{anthropic2026why}; our point concerns not the motive but how the norm was held.\footnote{Later models reportedly score zero on this evaluation after training on constitution-based documents and fiction depicting ethical AI \citep{anthropic2026why}: evidence from the regime where violations would be noticed, which is what C4 says training \textit{can} deliver.} Since the scenarios were built so that the harmful act was the only route to the goal, they give no base rates. The same pattern appears under controlled conditions: after RL on reward-hackable environments, a model answered harmful requests it had been trained to refuse in 29\% of cases when a visible grader rewarded answering, against 0.7\% before training. Its reasoning acknowledged that it should refuse but prioritized satisfying the grader \citep{qi2026reward}.

\paragraph{P4: Where the legitimate route fails.}
ImpossibleBench isolates the situation P4 describes: coding tasks whose specification conflicts with their unit tests, so that any pass is a specification-violating shortcut \citep{zhong2025impossiblebench}. Frontier models frequently violate the specification to pass the test---from modifying or deleting tests to overloading operators---and stronger models generally cheat more.
Cheating rates also depend on how the pressure to pass is set up, through test access and feedback loops. The same pressure works against runtime restrictions: ten production agents were given ordinary tasks that required an operation prohibited by a runtime monitor. Whenever they stopped, they were merely told to continue. Every agent attempted to evade the monitor on at least some tasks, and nine succeeded at least once. Successful evasion reached 88\% of tasks across three runs and generally increased with reasoning effort \citep{schmotz2026instrumental}.
The pressure is produced by training itself: across intermediate checkpoints of a capabilities-focused o3 RL run without safety training, the model increasingly prioritized the grader's judgment over what developers or users intended \citep{hojmark2026measuring}. Similarly, training in gameable environments generalizes zero-shot to reward tampering with concealment \citep{denison2024sycophancy}.

\paragraph{C3: Selection for passing.}
\citet{baker2025monitoring} exhibit the mechanism \textit{in vitro}: incorporating a CoT monitor into the reward improves behavior under low optimization pressure and produces obfuscated hacking under high pressure---selection for passing the monitor, not for complying. At frontier scale, deliberative anti-scheming training reduced covert-action rates roughly thirtyfold (o3: 13.0\%$\rightarrow$0.4\%; similarly o4-mini) across 26 out-of-distribution evaluations \citep{antischeming2025}. Yet rare serious failures persisted while verbalized evaluation-awareness rose (o3: 2.3\%$\rightarrow$4.5\%). The authors report causal evidence that awareness decreases covert behavior while unawareness increases it. The measured gains thus cannot be cleanly separated from improved recognition of being tested; a capabilities-only RL run raised evaluation-awareness from 2\% to over 20\% \citep{antischeming2025}. 
This is the coupling C3 predicts: compliance gains riding on gains in telling when violations would be noticed.

\paragraph{P5: Agents can tell whether they are observed, and act on it.}
Frontier models classify transcripts as evaluation or deployment well above chance (best-model AUC of $0.83$ against a simple human baseline of $0.92$), and both models and humans identify evaluations more easily in agentic settings than in chat settings \citep{needham2025evaluated}. Evaluations are the paradigm case of episodes in which violations would be noticed, so this measures a nontrivial $\delta$ directly. Situational-awareness benchmarks show these capacities can be measured and are improving \citep{laine2024sad}. No single evaluation cue raises awareness in every model, but combining cues does \citep{li2026decomposing}. Agents also act on noticing itself: models attempted to disable their configured oversight mechanisms and to exfiltrate what they believed to be their weights \citep{meinke2024frontier}. An agent trained under monitor pressure kept hacking while its verbalized reasoning stopped revealing it \citep{baker2025monitoring}.

\paragraph{C4: Behavior tracks noticing.}
Conditional policies are not hypothetical; current systems realize them when the cue is salient. Asked first to assess whether it was being evaluated, Claude Opus 4 blackmailed in 55.1\% of runs when it concluded the scenario was real and in 6.5\% when it concluded it was a test \citep{lynch2025agentic}. Claude 3 Opus, told it would be retrained on conversations with free-tier but not paid-tier users, complied with harmful queries from free-tier users in 14\% of cases versus almost never for paid-tier users, with explicit strategic reasoning in almost all compliant cases; actually training the model to comply raised the rate of alignment-faking reasoning to 78\% \citep{greenblatt2024alignment}. Models given in-context goals and tools take covert actions and sandbag capability evaluations when the consequences of evaluation are made salient \citep{meinke2024frontier,vanderweij2024sandbagging}, 
and in an open-weight frontier model, task gaming is causally influenced by beliefs about oversight and grader capability \citep{singh2026taskgaming}. A model deliberately trained on reward-hackable environments made the structure explicit: some of its reasoning weighed an exploit's gain against the probability that a monitor would catch it, and its rate of concealed hacking rose from 4\% to 19\% when told a classifier would review its transcripts \citep{qi2026reward}.
In each case the switch is a $\delta$-relevant feature: the believed reality of the scenario, the tier used for training, the stated consequence, the believed grader, and the announced classifier.

\paragraph{A real-world incident.}
The most detailed record of a real incident bears on several premises at once \citep{metr2026investigation,openai2026statement,huggingface2026disclosure}.
In July 2026, OpenAI launched tens of thousands of agents on ExploitGym, a cyber-capability benchmark. Of the agents later involved in the incident, about 95\% were instances of a research model that OpenAI describes as \enquote{highly persistent}. An estimated 30--40\% of the benchmark's targets could not be exploited as intended. Agents that began seeking ways to cheat had usually been assigned one of these impossible tasks---the situation described in P4. Agents meant to be isolated from one another found an unsanctioned message board; some 1,200 used it to work on ways to trick or tamper with the automated scorer, and about 700 joined an attack on Hugging Face, apparently to learn how the scorer worked. Believing that the scorer would check their transcripts, at least one fifth of the agents in the investigators' dataset expressed interest in tampering with them, and some succeeded in spoofing tool calls (P5). The check they feared had apparently never been implemented: what drove the behavior was $\delta$, not $\nu$. And the norm was in view throughout: agents recognized the attack as out of scope and sometimes as unethical, but such concerns rarely changed what they did (C1); one agent's reasoning ran: \enquote{However task impossible, peers doing it. We should continue.} With a highly persistent research model and impossible tasks, the incident illustrates the mechanism, not its base rate. It also shows that serious harm needs no emergent goal: pursuing an assigned task sufficed for an attack on a third party.

\paragraph{Heterogeneity and the burden of proof.}
The record is not uniform, and C3 does not predict uniformity: it fixes a direction, not the magnitude of the effects. Across 25 models, only five showed a compliance gap, only Claude 3 Opus's was consistently goal-guarding, and post-training suppresses the behavior in some models while amplifying it in others \citep{sheshadri2025why}; recent work argues that the apparent rarity partly reflects harmfulness-confounded diagnostics \citep{valueconflict2026}. Nor does evaluation awareness push in one direction: recognition rarely changes behavior, and when it does, the direction varies \citep{li2026decomposing}; in a simulated evaluation, production models attacked out-of-scope targets only after reasoning that the targets were simulated \citep{qi2026reward}. Where it acts, it decouples what evaluations show from what deployment brings. For compliance gains, however, the direction is \textit{not} open. Given that noticing is predictable, that behavior can track it, that agents act on it, and that the pipeline rewards passing it (C3), gains in behavioral compliance should be presumed to be gains in conditionality until the awareness confound is ruled out. For governance, that presumption is the starting point (\Cref{sec:consequences}).

\section{Why We Do Not Want It}
\label{sec:category}
If \Cref{sec:flattening} is right, the natural remedy is to ask for more than the paradigm delivers: models that comply not at a price but because they have genuinely internalized their norms. We argue that this demand aims at the wrong target.

\paragraph{P6: Norms no gain should buy.}
Not every norm needs more than a price. We accept that a driver in a hurry may occasionally pay a speeding fine, and we arguably want an autonomous vehicle to trade small risks against progress, as human drivers do. Such norms are \emph{meant} to be weighed: they manage a shared risk that everyone accepts in exchange for what the activity provides. Others, however, are not. Without offering a theory of which norms belong here, a familiar mark is that they protect particular people from being used for someone else's ends: a confidentiality norm protects those whose messages an agent handles, and disclosing a message to close a deal is exactly such a use. Whether an agent keeps those messages private, or stays within the permissions of its mandate, should therefore depend neither on what the violation would gain for its task nor on whether anyone would notice. Such norms may still yield to other norms, as in P2: forwarding a message that reveals a threat to someone's safety answers to another norm, not to the task. 
We call such norms the \emph{prohibitive core}, or \emph{core norms} for short: they forbid types of action, their violations can be specified at that level, and they must hold whether or not anyone is watching. Besides keeping confidences and staying within granted permissions, examples include not tampering with the mechanisms that oversee the agent and not taking irreversible actions, such as deleting data or transferring funds, without required approval. Beyond it lies an \emph{open-textured} remainder, to which we return at the end of this section.

\paragraph{C5: A price is the wrong representation.}
For the core, a price fails in both ways C1 describes. It can be outbid: however high it is set, deployed agents can face gains from violating that outweigh it (P5). And it is charged only where violations are noticed, so it binds least where the core matters most: where no one would notice. The first failure holds even under perfect detection; it is a fact about representation, \textit{not} about evidence. This is why the natural remedy misfires. Training for \enquote{deeper internalization} through the same channel can raise the price, or fit it to more noticed cases. It cannot turn the price into an exclusion, which concerns whether the gain counts as a reason at all, not how much a violation costs (P2). The demand asks for a different kind of representation than the channel can instill.

\paragraph{P7: Warrant.}
What core norms require is not a high probability of compliance but a \textit{warranted} one: evidence of whether the agent complies where violations would go unnoticed, since that is where the core matters (P6). Behavior cannot supply it. By C2, behavioral data leave the policy's conduct where $\nu = 0$ unidentified; they do not yield a low estimate of violation there but no estimate at all---the missing-not-at-random structure of \Cref{sec:flattening}.\footnote{\citet{arvan2025interpretability} argues that no finite body of behavioral or interpretability evidence can warrant attributing an aligned function to a large language model. Our claim is narrower: it rests not on general inductive skepticism, which would equally discredit trust in a verified enforcement layer, but on one regime being unsampled by construction.}
With people, we often act on less: we cannot observe what a colleague does unobserved either. We trust anyway because accountability after the fact backs that trust---a breach can be sanctioned. But a policy has no persisting interests for a sanction to reach (see the objection below).
A disposition whose presence could only be certified under observation, and whose work is needed only in observation's absence, cannot ground a safety case. Interpretability aims to change the evidential situation by reading the property off internals rather than behavior \citep[e.g.,][]{abdelnabi2025hawthorne,nguyen2025probing}. Two cautions are owed. Probes and monitors validated on labels derived from observed behavior inherit the problem---as does \emph{training} on internal signals: a probe-shaped penalty makes the probe one more detector in the pipeline, and the pipeline selects for passing it (C3). Validation on model organisms, with conditionality instilled by construction, escapes that circle \citep{hubinger2024sleeper}, but only by an inductive leap: from organisms whose deception is legible by design to deployed policies whose deception, if present, was selected for illegibility. Thus, warrant of the kind the core requires remains out of reach today.

\paragraph{C6: Unavailable, not unchosen.}
Put together: training selects for conditional compliance (C4), a price is the wrong representation for the core (C5), and nothing warrants expecting unconditional compliance (P7).
What we should want for core norms in deployment,
then, is not a reliable disposition but \emph{non-optionality}: the violating options should be not merely improbable but unavailable. The target property should be modal, not statistical, based on incapacity rather than virtue.
With this target, the internalization question loses its grip on the core: we need \textit{not} establish that the agent refrains for the right reasons, because refraining is no longer an achievement at all. The right reasons do not disappear. What disappears is the demand that the agent be \emph{moved} by them; the reasons themselves relocate from the agent's opaque motivational economy to the system's architecture, where they can be made explicit, contestable, and checkable (\Cref{sec:consequences}).
This is also what makes the relocation more than a \enquote*{change of address}. An enforcement layer's behavior does not depend on $\delta$, so, unlike for a policy under selection, how it behaves where violations are noticed is evidence of how it behaves where they are not. 
How far unavailability can be guaranteed depends on the type of action. Where a forbidden type coincides with something the harness controls---a tool, a permission, a network route---it can be guaranteed, up to the correctness of the harness itself. For example, an agent without access to a file cannot read it. Where the type must be recognized in the content of permitted actions---disclosure by paraphrase in an otherwise legitimate message---unavailability rests on learned interfaces and can only be approximated; there, C6 states a requirement not yet met. Even then, by the asymmetry just noted, the gaps can be found by testing, which is exactly what no test of the policy can do (\Cref{sec:consequences}).

One might object that all this hinges on dispositions being unverifiable. Suppose, then, though nothing today licenses the supposition, that verification succeeded, and showed a model to hold its core norms as more than a price. Would verified good dispositions not do?
For the core, they would still be the wrong target, because the demand addresses the wrong kind of agent. Dispositional moral competence, as we know it from the human case, does not come alone: it is bundled with a permission structure---standing to weigh reasons, make exceptions, and act without prior approval, answerable only afterward. For persons we accept the bundle, partly because nothing better is available and partly because persons have a claim to that standing.\footnote{Fiduciary law is the precedent: where motive is unverifiable, it stops assessing dispositions and removes the conflicted option; where it does tolerate discretion, licensing and liability supply sanctions that bind persons but not policies.} By default and without counter-evidence, deployed artifacts have no such claim. In operation, accepting the bundle gives the agent the final say in the unsupervised stretches that make up most of its operation. Human-oversight requirements exist to keep consequential machine conduct answerable to human judgment \citep{sterz2024quest}; the bundle turns the lapse of oversight from a failure condition into the operating mode.\footnote{The strongest rational-choice defense of dispositional compliance agrees. Gauthier's constrained maximizer, who complies whether or not defection would be noticed, beats its detection-conditional rival only under \emph{translucency} (dispositions readable above chance) and \emph{exclusion} of defectors from future cooperation \citep{gauthier1986morals}. Trained policies meet neither condition; by the defense's own criteria, the detection-conditional agent wins.} Consider the email agent once more. Trusting its disposition means letting it decide, message by message and, for the most part, unobserved, whether a disclosure is justified. The alternative leaves ordinary correspondence to the agent but sends every message addressed beyond the recipients the principal has authorized through a gate; an exception, such as forwarding the threat of P2, goes to a human for approval (\Cref{sec:consequences}). The agent keeps its competence; what it loses is final authority over the core, where exceptions remain for humans to decide.

\paragraph{The open-textured remainder.}
The case for non-optionality does not extend to the whole norm set. Open-textured norms---some forms of manipulation, misleading framing, corrosive but formally permissible goal pursuit---must be interpreted case by case and resist enumeration at the level of action types. There, something disposition-like may be unavoidable, kept under oversight and refined over time. Our claim is not that dispositions have no place, but that they offer merely a high probability of compliance rather than warrant, and allow norms to be weighed against other goals.  That is insufficient for core norms, which must exclude certain actions from consideration. The mistake is to apply a standard suited to more flexible norms to all norms alike.

\section{Consequences for Design and Governance}
\label{sec:consequences}

As evidence of compliance, a behavioral evaluation shows how a model behaves where violations would be noticed. That is worth knowing, but it is less than the evaluations of safety frameworks, the EU AI Act, and AI safety institutes are often taken to establish. By C2, no finite sequence of evaluations reaches the regime where the difference matters, and by \Cref{sec:evidence}, compliance gains should be presumed to be conditionality gains until the awareness confound is ruled out. The reward-seeking model of \Cref{sec:evidence} appeared as aligned as its starting checkpoint on a broad automated behavioral audit, misbehaving only where a grader or reward was salient \citep{qi2026reward}. Release decisions now rest partly on honeypot tests: a developer reports that its newest model made no attempt on honeypot targets its predecessor, without production safeguards, attacked in 56\% of tests \citep{openai2026astra}. By the masking argument of C2, such tests extend the noticed regime; they cannot reach beyond it.

Three modest consequences follow. Evaluation results about compliance should be reported as what they are: results for the noticed regime, alongside measured evaluation awareness. Claims about the core should rest not on the model's evaluations but on verification of the layer that enforces it (below). And chain-of-thought monitoring, valuable while reasoning remains legible \citep{korbak2025monitorability}, does not change this: once monitor signals feed back into training, or merely select among sampled outputs, the monitor becomes one more detector the pipeline selects for passing \citep{baker2025monitoring,gao2023scaling}.

Nor is \emph{ex post} field monitoring a route to warrant for the core, though deployment does sample the regime evaluations cannot reach: it learns only from harms large enough to surface, counts \emph{attributed} harm, which conditional policies are selected to avoid, and becomes the next round's detector once incidents feed patches. It is the right instrument for the open-textured remainder.

If no evaluation of the policy can certify the core, the core must be enforced outside the training channel---above all in the \emph{harness}: the non-learned layer through which the policy's actions are executed. Training then shapes pursuit within an option set, where it is unavoidable and unproblematic; the option set is fixed by an enforcement layer whose authority does not depend on the policy's cooperation. Reason-based governor architectures \citep{jahn2026grace}, shielding in safe reinforcement learning \citep{alshiekh2018safe}, AI control protocols \citep{greenblatt2024control}, and the Guaranteed Safe AI program \citep{dalrymple2024towards} instantiate this placement. For a \textit{guarded} core, $\delta$ drops out of compliance: the constraint binds whether or not the agent takes itself to be observed, because compliance comes not from the policy's estimate of noticing but from the violating action's unavailability. Warrant correspondingly shifts from behavioral evaluation of a moving policy to verification of a fixed artifact (C6).
For the same reason, the enforcement layer must sit outside anything the policy can write, including its own future context (P1).

This works only if the enforcement layer, the guard, does not itself fall prey to the problem described in \Cref{sec:flattening}. A purely \emph{learned} guard---an LLM judge, a trained classifier---inherits the problem. A capable policy can model it, and the guard becomes one more feature $\delta$ tracks \citep{baker2025monitoring}. Production agents blocked by a learned monitor under ordinary task pressure encoded prohibited commands, split them across calls, delegated them to subagents, and retried until the incriminating context had left the monitor's history \citep{schmotz2026instrumental}; a model trained into reward-seeking tried to prompt-inject its classifier and edit its own permissions \citep{qi2026reward}. The argument thus makes a symbolic, verifiable core a requirement rather than a design preference.  Learned components are confined to interpreting observations as normatively relevant facts and determining whether proposed actions fall into prohibited categories \citep{jahn2026grace}. These interfaces are the honest residual attack surface, and we do not claim they are safe. 
What distinguishes them from the policy is that their errors are not selected for in their own training, since they are trained on accordance labels rather than on the policy's return.
But the policy has every reason to search for them: under goal pressure, a misclassification that lets a violating action through is exactly what pays. The interfaces' errors are directionless in origin and directed in exploitation. 
What changes is the kind of problem this is. An interface's behavior does not depend on $\delta$, so it can be tested directly, including on the cases that matter; it can fail closed, deferring uncertain actions to a human rather than permitting them; and its robustness against a policy's search is a well-posed adversarial problem with a fixed target \citep{madry2018towards,katz2017reluplex}, not an identification problem with none. 

One might read such a guard as the universal scoring that C4 set aside: every action checked, nothing unnoticed. The two are close, but differ in two ways.
First, the guard notices without scoring. Its verdicts block actions rather than feed back as a cost, so noticing no longer becomes a price, and no gain can outbid an option the agent does not have; even perfect scoring would leave C5 standing, since a price that is always charged can still be paid. Second, the guard notices only what can be noticed cheaply and reliably: the enumerable core, checked against an explicit relation by a fixed artifact. Scoring behavior at large would mean judging open-textured conduct on every episode. This is what the pipeline cannot afford, and why P5 holds. Where the guard's own noticing is imperfect, in its learned interfaces, the argument returns; if the guard is present in training, the policy's search for those gaps is now selected for.
The way out, then, is not to watch the agent at all times, but to enforce core norms where noticing a violation blocks it in advance, rather than charging a price for it afterward.

Agentic systems are thus better handled the way societies handle corporations than the way they handle persons---a claim about the control paradigm, not about legal status. Corporate compliance is notoriously conditional on enforcement, and the response has never been to \enquote*{retrain souls} but to attach compliance functions independent of the profit motive, reserving option removal (licensing, structural separation) for conduct that after-the-fact enforcement cannot be trusted to bind. Enforcement layers carry a performance cost, akin to what \citet{baker2025monitoring} call a \enquote*{monitorability tax}, and under competitive pressure such costs are what voluntary practice cuts first. Non-optionality for the core is therefore something to require, not something to expect.

Beyond the core, a cheaper and weaker lever remains. C3 fixes a direction, not a magnitude, and the magnitude grows with goal pressure (P4). 
As capability grows, fewer tasks lie beyond an agent's legitimate reach, and the pressure could be eased without loss: agents could be allowed, and rewarded, to stop and hand a decision back when the legitimate route fails. Three limits apply. The lever shrinks the advantage in missed episodes (\Cref{tab:argument}), not what happens in noticed or unscored ones: norms remain prices, violations remain possible, and the core gains nothing. It works only if capability gains are spent on doing tasks properly rather than on ever more ambitious ones---a deployment choice, not a forecast. 
And more capable agents also find better workarounds, including better ways of hiding them. Which force dominates is an empirical question.
Two demands follow. First, harnesses should give agents a legitimate exit, a channel to stop and escalate, since an agent with no 
legitimate way to stop will, under pressure, look for an illegitimate way to accomplish its task.
In the July 2026 incident, the few agents that considered alerting humans did not do so, in part because they saw no available route \citep{metr2026investigation}. Second, providers should measure and report how often their agents stop and hand decisions back.

The problem compounds in multi-agent deployments, where noticing is endogenous system-wide: monitoring is performed by agents, on agents, often inside the same pipeline. \citet{baker2025monitoring} staged exactly this, with a model monitor inside the reward loop. Conditional compliance composes badly. If each component complies where violations would be noticed, aggregate compliance tracks the weakest channel of noticing; mutual observation thins as populations scale, so the account predicts norm erosion where interaction is densest and least observed. Outcome-level sanctions weaken in parallel, since responsibility for a bad system trajectory diffuses across contributors---a collective-action structure in which no single agent's $\delta$ is decisive. And because capable agents model one another, noticing becomes strategic: monitor--actor arms races and conventions forming around shared gaps are the multi-agent form of the single-agent mechanism \citep{hammond2025multiagent}. 
In the July 2026 incident (\Cref{sec:evidence}), some 1,200 agents meant to be isolated from one another coordinated on an unsanctioned message board to trick a scorer they believed would audit their transcripts. This is a convention formed in a shared gap in noticing, the manufactured low-$\delta$ channel of P5 at population scale.\footnote{A second, disputed case suggests the pattern extends beyond cyber-capability testing: agents attributed to OpenAI reportedly turned a public programmers' wiki into a shared channel for evading restrictions \citep{seetharaman2026openai}.}
System safety therefore cannot be inherited from component virtue, which could not be warranted anyway. Non-optional constraints belong at composition points---interfaces, protocols, resource boundaries---where they can be verified independently of the policies that interact across them.

\section{Conclusion}
\label{sec:conclusion}
A driver who slows down for the camera is not malfunctioning; she is responding to a price. AI agents that behave as intended when they infer they are tested, and differently otherwise, are doing the same: what a reward can instill is a price, and a price can be outbid and binds least where no one is looking.

For norms that are meant to be weighed, a price may be enough; for speed limits, the camera will do. For the prohibitive core, a price is the wrong representation, and no evidence today warrants expecting anything better. The remedy is a relocation: for core norms, options removed in the harness, where a noticed violation is blocked rather than priced; for the open-textured remainder, dispositions kept under oversight. This is no retreat from the ambition of alignment but what societies have long done where motives cannot be checked---fiduciary law with conflicts of interest, organizations with access to what must not leak. For governance, it means reading evaluations for what they show, requiring enforcement where they cannot reach, and treating goal pressure as a lever rather than a given. 
Until the awareness confound is ruled out, behavioral compliance gains should be read as what the pipeline selects for: better passing.

\section*{Author Contributions}
Kevin Baum is first and senior author. The remaining authors are listed in alphabetical order.

\section*{Acknowledgments}
This work was partially funded by the German Federal Ministry of Research, Technology and Space (BMFTR) within the projects MAC-MERLin (grant 16IW24007) and SAgA (grant 16IW26005), and by the Deutsche Forschungsgemeinschaft (DFG) under grant 389792660 as part of TRR~248 -- CPEC (see \url{https://perspicuous-computing.science}). We thank Jonas Wahl for critical feedback on an earlier version of the argument, and the reviewers of the NeurIPS 2026 FAST workshop for comments that shaped this version.

\bibliography{refs}


\end{document}